\documentclass{article}[10pt]
\pdfoutput=1
\usepackage{jucs2e}
\usepackage{url}
\usepackage{graphicx}
\begin{document}

\title{Review-Level Sentiment Classification with Sentence-Level Polarity Correction}

\author{{\bfseries Sylvester Olubolu Orimaye\\
   (School of Information Technology,\\
Monash University Malaysia\\
   sylvester.orimaye@monash.edu)
   \and
   {\bfseries Saadat M. Alhashmi}\\
   (Department of Management Information Systems,\\
University of Sharjah,
\\ Sharjah, United Arab Emirates \\
   salhashmi@sharjah.ac.ae)\\
   \and
   {\bfseries Eu-Gene Siew}\\
   (School of Business,\\
Monash University Malaysia\\
   siew.eu-gene@monash.edu)\\
   \and
   {\bfseries Sang Jung Kang}\\
   (School of Information Technology,\\
Monash University Malaysia\\
   sjkan2@student.monash.edu)\\
}}

\maketitle

\begin{abstract}
We propose an effective technique to solving review-level sentiment classification problem by using sentence-level polarity correction. Our polarity correction technique takes into account the consistency of the polarities (positive and negative) of sentences within each product review before performing the actual machine learning task. While sentences with inconsistent polarities are removed, sentences with consistent polarities are used to learn state-of-the-art classifiers. The technique achieved better results on different types of products reviews and outperforms baseline models without the correction technique. Experimental results show an average of 82\% F-measure on four different product review domains.
\end{abstract} 


\section{Introduction}
\label{section:introduction}

Sentiment classification has attracted a number of research studies in the past decade. The most prominent in the literature is Pang et al,\cite{pang:2002} which employed supervised machine learning techniques to classify positive and negative sentiments in movie reviews. The significance of that work influenced the research community and created different research directions within the field of sentiment analysis and opinion mining.\cite{vechtomova:2010,liu:2012,cambria:2013} Practical benefits also emerged as a result of automatic recommendation of movies and products by using the sentiments expressed in the related review.\cite{john_blitzer:2007,liu:2012,cambria:2013} This is also applicable to business intelligence applications which rely on customers' reviews to extract `satisfaction' patterns that may improve profitability.\cite{martinez:2014,vilares:2015} While the number of reviews has continued to grow, and sentiments are expressed in a subtle manner, it is important to develop more effective sentiment classification techniques that can correctly classify sentiments despite natural language ambiguities, which include the use of irony.\cite{mendel:2010,keshtkar:2012,reyes:2014,melero:2014} 

In this work, we classify sentiments expressed on individual product types by learning a language model classifier. We focus on online product reviews which contain individual product domains and express explicit sentiment polarities. For example, it is quite common that the opinion expressed in reviews are targeted at the specific products on which the reviews are written.\cite{vechtomova:2010,vilares:2015} This enables the reviewer to express a substantial level of sentiments on the particular product alone without necessarily splitting the opinions between different products. Also, in a review, sentiments are likely to be expressed on specific aspects of the particular product.\cite{jo:2011} For example, an iPad user may express positive sentiment about the `camera quality' of the device but expresses negative sentiment about the `audio quality' of the device. This provides useful and collaborative information on aspects of the product that need improvements.\cite{tsai:2013,poria:2013,roman:2015} 

The application of sentiment classification is important to the ordinary users of opinion mining and sentiment analysis systems.\cite{pang:2008,vechtomova:2010,liu:2012} This is because the different categories of sentiments (e.g. positive and negative) represent the actual stances of humans on a particular target (e.g. a product). A product manufacturer for example, can have an overview of how many people `like' and `dislike' the product by using the number of \emph{positive} and \emph{negative} reviews. Similarly, sentiment classification has been quite useful in finance industries, especially for stock market prediction.\cite{devitt:2007,brabazon:2008,fortuny:2014} 

Sentiment classification on product reviews can be challenging,\cite{pang:2008,liu:2012,cambria:2013} which is why it is still a very active area of research. More importantly, sentiments expressed in each product review sometimes include ambiguous and unexpected sentences, \cite{li:2010extraction} and are often alternated between the two different positive and negative polarities. This causes inconsistencies in the sentiments expressed and consequentially leading to the mis-classification of the review document.\cite{pang:2002,pang:2008,liu:2012} As such, the bag-of-words approach is not sufficient alone.\cite{liu:2012,cambria:2013} We emphasize that most negative reviews contain positive sentences and often express negative sentiments by using just a few negative sentences.\cite{jia:2009} We show an example as follows: 

\vspace{0.2cm}
\begin{center}
\parbox[t]{3.5in}{\footnotesize 
\emph{I bought myself one of these and used it minimally and was happy} (POSITIVE)
\newline \emph{I am using my old 15 year old Oster}  (NEGATIVE)
\newline \emph{Also to my surprise it is doing a better job} (POSITIVE)
\newline \emph{Just not as pretty} (NEGATIVE)
\newline \emph{I have KA stand mixer, hand blender, food processors large and small...} (OBJECTIVE)
\newline \emph{Will buy other KA but not this again} (NEGATIVE)
 }
\end{center}
\vspace{0.2cm}

\noindent The above problem often degrades the accuracy of sentiment classifiers as many review documents get mis-classified to the opposite category. This is regarded as \emph{false positives} and \emph{false negatives} as the case may be.
 
While the above problem is non-trivial, we propose a polarity correction technique that extracts sentences with consistent polarities in a review. Our correction technique includes three separate steps. First, we perform training set correction by training a `na\"{\i}ve' sentence-level polarity classifier to identify \emph{false negatives} in both positive and negative categories. We then combine the \emph{true positives} sentences and the \emph{false negative} sentences of the two opposite categories to form a new training set for each category. Second, we propose a sentence-level polarity correction algorithm to identify consistent polarities in each review, while discarding sentences with inconsistent polarities. Finally, we learn different Machine Learning algorithms to perform the sentiment classification task.

The above steps were performed on four different Amazon product review domains and improved the accuracy of sentiment classification of the reviews over a baseline technique and give comparable performance with standard biagram, bag-of-words, and unigram techniques. In terms of F-measure, the technique achieve an average of 82\% on all the product review domains.  

The rest of this paper is organized as follows. We discuss related research work in Section \ref{section:related-work}. In Section \ref{section:training-set-correction}, we propose the training set correction technique for sentiment classification task.  Section \ref{section:polarity-correction} describes the sentence-level polarity correction technique and the corresponding algorithm. Our machine learning experiments and results are presented in Section \ref{section:experiment-result}. Finally, Section \ref{section:conclusion} presents conclusions and future work.

\section{Related Work}
\label{section:related-work}

Pang and Lee,\cite{pang:2004} proposed a subjectivity summarization technique that is based on minimum cuts to classify sentiment polarities in IMDb movie reviews. The intuition is to identify and extract subjective portions of the review document using minimum cuts in graphs. The minimum cut approach takes into consideration, the pairwise proximity information via graph cuts that partitions sentences which are likely to be in the same class. For example, a strongly subjective sentence might have lexical dependencies on its preceding or next sentence. Thus Pang and Lee,\cite{pang:2004} showed that minimum cuts in graph put such sentences in the same class. In the end, the identified subjective portions as a result of the minimum graph cuts are then classified as either negative or positive polarity. This approach showed significant improvement from 82.8\% to 86.4\% with just 60\% subjective portion of the documents. 

In our work, we introduce additional steps by not only extracting subjective sentences. Instead, we extract subjective sentences with consistent sentiment polarities. We then discard other subjective sentences with inconsistent sentiment polarities that may contribute noise and reduce the performance of the sentiment classifier. Thus, contrary to Pang and Lee,\cite{pang:2004} our work has the ability to effectively learn sentiments by identifying the likely subjective sentences with consistent sentiments. Again, we emphasize that some subjective sentences may not necessarily express sentiments towards the subject matter.\cite{liu:2012,cambria:2013} Consider, for example, the following excerpt from a `positive-labelled' movie review:

\vspace{0.2cm}
\begin{center}
\parbox[t]{3.5in}{\footnotesize \emph{`$ ^{1}$real life, however, consists of long stretches of boredom with a few dramatic moments and characters who stand around, think thoughts and do nothing, or come and go before events are resolved. $ ^{2}$Spielberg gives us a visually spicy and historically accurate real life story. $ ^{3}$You will like it.'}}
\end{center}
\vspace{0.2cm}

\noindent In the above excerpt, sentence 1 is a \emph{subjective} sentence which does not contribute to the sentiment on the movie. Explicit sentiments are expressed in sentence 2 and 3. We propose that discarding sentences such as sentence 1 from reviews is likely to improve the accuracy of a sentiment classifier. 

Similarly, Wilson et al,\cite{wilson:2005a} used instances of polar words to detect contextual polarity of phrases from the MPQA corpus. Each phrase detected is verified to be either \emph{polar} or \emph{non-polar} phrase by using the presence of opinionated words from a polarity lexicon. Polar phrases are then processed further to detect their respective contextual polarities which can then be used to train machine learning techniques. Identifying the polarity of phrase-level expression is a challenge in sentiment analysis.\cite{liu:2012} Earlier in Section \ref{section:introduction}, we have illustrated some example sentences to that effect. For clarity, consider the sentence \emph{`I am \textbf{not} saying the picture quality of the camera is \textbf{not good}'}. In this sentence, the presence of the negation word \emph{\textbf{`not'}} does not represent `negative' polarity of the sentence in context. In fact it emphasizes a `desired state' that the `picture quality' of the camera entity is `good'. However, without effective contextual polarity detection, such sentences could be easily classified as `negative' by ordinary machine learning techniques. To this extent, Wilson et al,\cite{wilson:2005a} performed manual annotation of contextual polarities in the MPQA corpus to train a classifier with a combination of ten features resulting to 65.7\% accuracy giving room for more improvement.

Choi and Cardie,\cite{choi:2008} proposed a \emph{compositional semantics} approach to learn the polarity of sentiments from the sub-sentential level of opinionated expressions. The compositional semantic approach breaks the lexical constituents of an expression into different semantic components. Thus, the work used \emph{content word negators} (e.g. sceptic, disbelief) to identify the sentiment polarities from the different semantic components of the expression. Content word negators are \emph{negation} words other than \emph{function} words such as \emph{not}, \emph{but}, \emph{never} and so on. Identified sentiment polarities are then combined using a set of heuristic rules to form an overall sentiment polarity feature which can then be used to train machine learning techniques. Interestingly, on the Multi-Perspective Question Answering (MPQA) corpus created by Wiebe et al,\cite{wiebe:2005} this combination yielded a performance of 90.7\% over the 89.1\% performance of ordinary classifier (e.g. using bag-of-words). 

The performance achieved by Choi and Cardie,\cite{choi:2008} is understandable given that the MPQA corpus contains well `structured' news articles which are mostly well written on certain topics. Moreover, sentences or expressions which are contained in news articles are most likely to express \emph{sequential sentiments} for a reasonable classification performance.\cite{devitt:2007,bautin:2010,lee:2010a} For example, it is more likely that a negative news `event' such as \emph{`Disaster unfolds as Tsunami rocks Japan'} will attract `persistent' negative expressions and sentiments in news articles. In contrast, sentiment classification on product reviews is more challenging as there is often inconsistent or mixed sentiment polarities in the reviews. We have illustrated an example to that effect in Section \ref{section:introduction}. It would be interesting to know the performance of the heuristics used by  Choi and Cardie,\cite{choi:2008} on standard product review datasets such as Amazon online product review datasets. A detailed review of other sentiment classification techniques on review documents is provided in Tang et al.\cite{tang:2009} 

Our main contribution to the sentiment classification task is to do training set correction and further detect \emph{inter-sentence} polarity consistency that could improve a sentiment classifier. That is, given a review of $n-$sentences, we try to understand how the sentiment polarity varies from sentence 1 to sentence $n$.  We hypothesize that detecting \emph{consistent sentiment patterns} in reviews could improve a sentiment classifier without further sophisticated natural language techniques (e.g. using compositional semantics or linguistic dependencies).\cite{devitt:2007}

More importantly, we believe every sentence in the review may not necessarily contribute to the classification of the review to the appropriate class.\cite{pang:2004} We say that certain \emph{sequential sentences with consistent sentiment polarities} could be sufficient to represent and distinguish between the sentiment classes of a review. Representative features have been argued to be the key to effective classification technique.\cite{arel:2010,he:2011} We emphasize that our approach is promising and can be easily integrated by any sentiment classification system regardless of the sentiment detection technique employed.

\section{Training Set Correction}
\label{section:training-set-correction} 

Training set polarity correction has been largely ignored in sentiment classification tasks.\cite{martinez:2014} Earlier, we emphasized that a review document could contain both positive and negative sentences. Moreover, since reviewers often express sentiments on different aspects of products, it is probable that some aspects of the products will receive positive sentiments while others get negative sentiments.\cite{liu:2012} In a negative-labeled product review for example, it is more likely that negative sentiments will be expressed within the first few portion of the review and then followed by positive sentiments in the later portion of the review on some of the aspects of the product that gave some satisfactions.\cite{john_blitzer:2007,liu:2012} This could be because reviewers tend to emphasize on the negative aspects of a product than the positive aspects, and in some cases, both polarities are expressed alternately, which we will discuss in Section \ref{section:polarity-correction}. Thus, using such mixed sentiments in each category, for training a machine learning algorithm will only result to bias and reduce the accuracy of the classifier.\cite{pang:2004,calais:2011} 

As such, we propose a promising approach to reduce the bias in the training set by first learning a `na\"{\i}ve' sentence-level classifier on all sentences from both the positive and negative categories. A `na\"{\i}ve' classifier could be any classifier trained with surface-level features (e.g. unigram or bag-of-words),\cite{pang:2004,hang_cui:2006} without necessarily performing sophisticated features engineering since the final sentiment classifier will be constructed with more fine-grained features. \cite{maldonado:2014} For example, one could learn the popular Na\"{\i}ve Bayes classifier with only unigram features.\cite{pang:2004,tan:2009,ye:2009} It is also possible to use a more complexly constructed classifier at the expense of efficiency. Having said that, the `na\"{\i}ve' classifier is then used to also test the same sentences from both the positive and negative categories. The idea is to identify \emph{positive-labelled sentences} that will be classified as negative and \emph{negative-labelled sentences} that will be classified as positive. Having identified this, it is therefore imperative to correct the training set by combining the wrongly classified sentences to their original respective categories. That is, positive-labelled sentences that are classified as negative should be combined with the original negative sentences (in the negative category) and  negative-labelled sentences that are classified as positive should be combined with the original positive sentences (in the positive category). 

While this technique may result to a meta classification,\cite{xia:2013} we propose to include the technique as part of the training process of the final sentiment classifier. In addition, in order to minimize wrongly classified sentences, we implement the `na\"{\i}ve' classifier to maximize the \emph{Joint-Log-Probability} score of a given sentence belonging to either of positive or negative categories. This is because most ordinary classifiers maximize the conditional probability over all categories, which is at the expense of better accuracy.\cite{lowd:2005} We compute the \emph{Joint-Log-Probability} as follows:

\begin{equation}
P(S,C) = \log_{2} P(S|C) + \log_{2} P(C)
\end{equation}

\begin{equation}
P_{c} = argmax_{c \in C} P(S,C)
\end{equation} 

where $P(S,C)$ is the probability of a sentence given a class, $P_{c}$ is the probability of the sentence belonging to either a `positive' category $c$ or a `negative' category $c$ and $P(C)$ is a multivariate distribution on the positive and negative categories.

\section{Sentence-Level Polarity Correction}
\label{section:polarity-correction} 

Following the training set correction in Section \ref{section:training-set-correction}, we propose the sentence-level polarity correction to further reduce mis-classification in both `training' and `testing' sets. More importantly, because the bag-of-words approach has seldom improve the accuracy of a sentiment classifier,\cite{liu:2012,cambria:2013} a sentence-level approach could give better improvement since most sentiments are expressed at sentence-level anyway.\cite{tan:2011} However, we have indicated in Section \ref{section:training-set-correction} that many review documents have the tendency to contain both positive and negative sentences, regardless of their individual categories (i.e. positive or negative). While the consistent sentence polarities of both categories might be helpful to the classification task, it would be better to remove sentences with \emph{outlier} polarities that cause inconsistencies by using a polarity correction approach.\cite{wilson:2009,li:2010,liu:2012} Note that we have motivated the inconsistency problem with an example in Section \ref{section:introduction}.

\begin{figure*}
\begin{center}
\includegraphics[width=35mm,height=45mm]{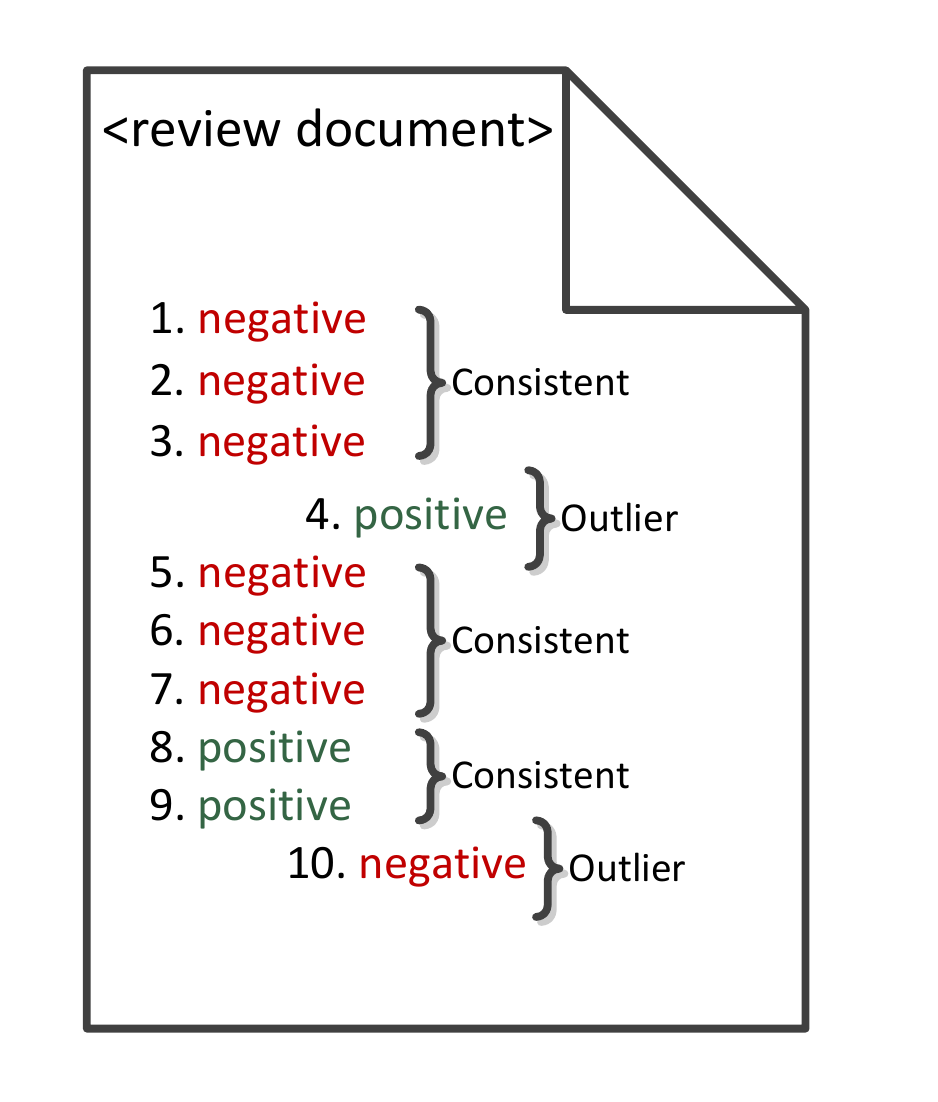}
\caption{An illustration of a review document with outliers, consistent, and inconsistent polarities.}
\label{fig:review-doc}
\end{center}
\end{figure*}

The idea of the sentence-level polarity correction is to remove inconsistent sentence polarities from each review. We observed that sentences with inconsistent polarity deviate from the previous consistent polarity. More often than not, the polarities of sentences in a given review are expressed consistently except for some outliers polarities.\cite{pang:2002,li:2010,liu:2012} As such, a given polarity is expressed consistently over a number of sentences and at a certain point deviate to the other polarity, and continues over a number of sentences alternately. Figure \ref{fig:review-doc} shows an illustration depicting a possible review with consistent polarities and inconsistent polarities (or outlier polarities).

Given a 10-sentence review, a reviewer has expressed negative sentiments with the first three sentences. This is followed by a single positive sentence on line 4. Lines 5 to 7 consist of another three negative sentences. Lines 8 to 9 expressed positive sentences. Finally, line 10 concluded with a negative sentence. Thus, we regard line 4 (positive sentence) and line 10 (negative sentence) as outlier polarities because there is no subsequent exact polarity after each of them. Our polarity correction algorithm removes such outliers, leaving only the consistent polarities. It is to be noted that at this stage, the algorithm is independent of a particular sentiment category (i.e. positive or negative). We consider exact subsequent polarities - either positive or negative - since a review is likely to contain both polarities as discussed earlier. Our intuition is that sentences with consistent polarity could better represent the overall sentiment expressed in a review document by providing a wider margin between the categories of the major consistent sentiment polarities.\cite{devitt:2007,jia:2009} Note that this technique is different from \emph{intra-sentence} polarity detection as studied in Li et al.\cite{li:2010} An additional thing we did was to performed negation tagging by tagging 1 to 3 words after a negation word in each sentence. In contrast to our baseline, the negation tagging showed some improvements in our correction technique.

Thus, given a review document with $n$-number of sentences $S_{1},...,S_{n}$, we classify each sentence with the `na\"{\i}ve' classifier and compare the polarity $\Phi_{s}$ of the first sentence with the polarity $\Phi_{s_{n} + 1}$ of the next sentence until $s_{n-1}$. Where $\Phi_{s}$ is the starting polarity, the polarity of the subsequent sentence $\Phi_{s_{n + 1}}$ is compared with the polarity of the \emph{prior} sentence $\Phi\lambda_{s_{n+1}}$. When $\Phi\lambda_{s_{n+1}}$ equals $\Phi_{s_{n + 1}}$, the sentence is stored into the consistent category, otherwise, the sentence is considered outlier. Note that we set a \emph{consistency} threshold by specifying a parameter $\theta$, which indicates the minimum number of subsequent and the same sentence polarities that must be considered consistent. As such, consistent sentence polarities that are lower than the $\theta$ value are ignored. 
	
\begin{figure*}
\begin{center}
\includegraphics[width=45mm,height=55mm]{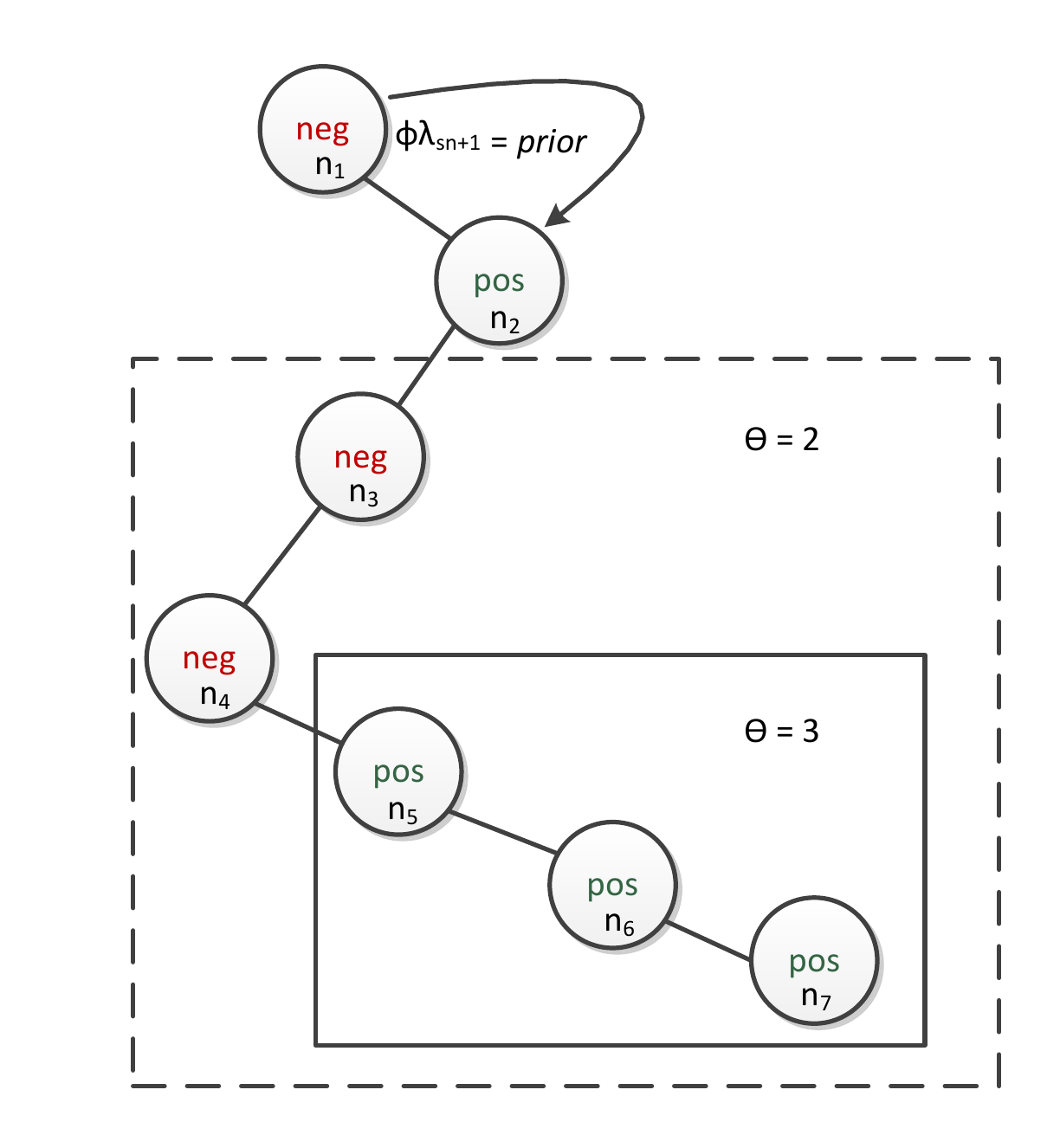}
\caption{Extracting consistent polarities with different $\theta$ thresholds.}
\label{fig:polarity-tree}
\end{center}
\end{figure*}

In our experiment, we set $\theta=2$ to simulate the default case. Our empirical observation shows that $\theta=2$ sufficiently captures consistent polarities for a sparse review document containing as low as 7 sentences. Figure \ref{fig:polarity-tree} shows how consistent polarities are extracted with different threshold $\theta$, where $\theta=2$ retrieves sentences $n_{3}$ to $n_{7}$ and $\theta=3$ retrieves only sentences $n_{5}$ to $n_{7}$. 

\section{Experiment and Results}
\label{section:experiment-result}  

We performed several experiments with our correction technique and compare between the performance on popular state-of-the-art classifiers with and without our polarity correction techniques. The classifiers comprises of the Sequential Minimum Optimization (SMO) variant of Support Vector Machines (SVM),\cite{platt:1998} and Na\"{\i}ve Bayes (NB) classifier.\cite{rish:2001} We used SVM and NB on the WEKA machine learning platform,\cite{hall:2009} with \emph{bag-of-words},\emph{unigram}, and word \emph{bigram} features. We did not include word \emph{trigram} features as both word unigram and word bigram features have been studied to improved sentiment classification tasks.\cite{pang:2002,pang:2004,liu:2012} We conducted \textbf{80\%}-\textbf{20\%} performance evaluation for comparison with the baselines on each dataset domain. 

For selecting the best parameters for the baseline algorithms, we performed hyperparameters search using Auto-Weka,\cite{thornton:2013} with cross-validation and the Sequential Model-based Algorithm Configuration (SMAC) optimization algorithm, which is an Bayesian optimization method proposed as part of Auto-Weka.\cite{thornton:2013} We performed the search by using the unigram features on the training set of each domain. This is because unigram features have shown robust performance in sentiment analysis.\cite{pang:2002,pang:2004,liu:2012} 

\begin{table*}[ht]
\footnotesize
	\begin{center}
	\begin{tabular}{p{3cm} p{9cm}}
  	\hline	
    \textbf{Model}    &  \textbf{Hyperparameters}\\
    \hline   
    SVM-beauty    &  -C 1.1989425641153333 -N 0 -K ``NormalizedPolyKernel -E 1.6144079568156302 -L"   \\
    SVM-books   &   -C 1.2918141993816825 -N 2 -K ``NormalizedPolyKernel -E 2.78637472738497"   \\
    SVM-kitchen   &   -C 1.2929645940353218 -N 2 -K ``Puk -S 9.028189222927269 -O 0.9952824838773323"   \\
    SVM-software   &   -C 1.1471978195519354 -N 2 -M -K ``NormalizedPolyKernel -E 1.7177045231155679 -L"   \\
    NB-all-domains    	&  -K (Kernel Estimator)   \\
    \hline
	\end{tabular}
	\end{center}
	\caption{Auto-Weka hyperparameters settings for SVM and NB on product domains with unigram features.}
\label{table:exp2}
	\end{table*}

\subsection{Dataset and Baseline}
\label{section:dataset-baseline} 

Our dataset is the multi-domain sentiment dataset constructed by Blitzer et al.\cite{john_blitzer:2007} The dataset was first used in year 2007 and consists of Amazon online product reviews from four different types of product domains\footnote{http://www.cs.jhu.edu/~mdredze/datasets/sentiment/}, which includes, \emph{beauty products},\emph{books}, \emph{software}, and \emph{kitchen}. Each product domain has \textbf{1000} positive reviews and \textbf{1000} negatives reviews, which were identified based on the customers' star ratings according to Blitzer et al.\cite{john_blitzer:2007} For each domain, we separated \textbf{800} documents per category as \emph{training set} and used the remaining \textbf{200} documents as \emph{unseen testing set}. We extracted the review text and performed sentence boundary identification by optimizing the output of the \emph{MedlineSentenceModel} available as part of the LingPipe library.\footnote{http://alias-i.com/lingpipe/docs/api/com/aliasi/sentences/MedlineSentenceModel.html} 

As our baseline, we implemented a sentence-level sentiment classifier using a technique similar to Pang and Lee,\cite{pang:2004} on the same dataset but without our correction technique. The baseline technique has worked very well in most sentiment classification tasks. The baseline work removes objective sentences from the training and testing documents by using an automatic \emph{subjectivity detector} component which uses subjective sentences only for sentence-level classification. 

\subsection{Evaluation}
\label{section:evaluation} 

We used three evaluation metrics comprising of \emph{precision}, \emph{recall}, and \emph{F-Measure} or \emph{F-1}. The \emph{precision} is calculated as $TP/(TP+FP)$, \emph{recall} as $TP/(TP+FN)$, and \emph{F-Measure} as  $(2 * precision * recall)/(precision + recall)$. Note that TP, TN, FP, and FN are defined as true positives, true negatives, false positives, and false negatives, respectively. All results are based on 95\% Confidence Interval.

\subsection{Results and Discussion}
\label{section:results-discussion} 

We present the results in Tables \ref{tab:beauty} - \ref{tab:kitchen}, where \emph{Model} is the type of classifier, \emph{Pr.} is the precision, \emph{Rc.} is the recall, and \emph{F-1} is the F-measure, respectively. We identify the models with our correction technique with `cor' after the model names. For example `SVM-Unigram-Cor' depicts a model using SVM with unigram features and our correction techniques.  Standard models are identified by the algorithm name and the feature used. Baseline models are identified with `Baseline'. In addition, we identify our best performing model above the baseline with ($\ast$) and comparable performance with standard models is identified with ($\dagger$). 

\begin{table*}\footnotesize
\centering
\makebox[0pt][c]{\parbox{1\textwidth}{%
    \begin{minipage}[t]{0.45\hsize}\centering
        \begin{tabular}{ llll }
           \hline
          \textbf{Model}  & \textbf{Pr.} & \textbf{Rc.} & \textbf{F-1} \\ \hline
SVM$-$Bigram$-$cor	& 	0.83 &	0.83 &	0.83\\
SVM$-$BOWS$-$cor	& 	0.83 &	0.83 &	0.83 \\
SVM$-$Unigram$-$cor	&  \textbf{0.84} &	\textbf{0.85}	& \textbf{0.84*} \\
SVM$-$Bigram		&	0.83 &	0.83 &	0.83 \\
SVM$-$BOWS		&  	0.85 & 	0.85 &	0.85$\dagger$ \\
SVM$-$Unigram		& 	0.83 &	0.83 &	0.83\\
SVM$-$Baseline	& 	0.76 &	0.76 &	0.76\\ \hline
NB$-$Bigram$-$cor	& 	0.83 &	0.79 &	0.80\\
NB$-$BOWS$-$cor		& 	\textbf{0.83} &	\textbf{0.81} &	\textbf{0.81*} \\
NB$-$Unigram$-$cor	& 	0.79 &	0.74 &	0.75 \\
NB$-$Bigram		&	0.86 &	0.86 &	0.86$\dagger$ \\
NB$-$BOWS			&	0.83 &	0.83 &	0.83 \\
NB$-$Unigram		& 	0.86 &	0.85 &	0.86\\
NB$-$Baseline		& 	0.75 &	0.75 &	0.75\\ \hline
        \end{tabular}
        \caption{Performance of unseen test sets on Beauty Reviews}
        \label{tab:beauty}
    \end{minipage}
    \hspace{\fill}
    \begin{minipage}[t]{0.45\hsize}\centering
        \begin{tabular}{ llll }
           \hline
          \textbf{Model}  & \textbf{Pr.} & \textbf{Rc.} &\textbf{F-1} \\ \hline
SVM$-$Bigram$-$cor & 	0.77 &	0.77 &	0.77\\
SVM$-$BOWS$-$cor	& 	\textbf{0.81} &	\textbf{0.81} &	\textbf{0.81*} \\
SVM$-$Unigram$-$cor	&  0.75 &	0.75 &	0.75\\
SVM$-$Bigram &	 0.76 &	0.75&	0.76 \\
SVM$-$BOWS &  0.78 & 0.78 &	0.78\\
SVM$-$Unigram & 	0.78 &	0.77 &	0.77\\
SVM$-$Baseline & 	0.68 &	0.68 &	0.68\\ \hline
NB$-$Bigram$-$cor & 	0.78 &	0.74 &	0.74\\
NB$-$BOWS$-$cor	& \textbf{0.84} &	\textbf{0.81} &	\textbf{0.81*} \\
NB$-$Unigram$-$cor	& 0.79 &	0.77 &	0.77 \\
NB$-$Bigram &	0.78 &	0.77 &	0.76 \\
NB$-$BOWS & 0.56 &  0.55 &	 0.53\\
NB$-$Unigram & 0.69 &	0.67 &	0.66\\ 
NB$-$Baseline & 0.69 &	0.67 &	0.64\\ \hline

        \end{tabular}
        \caption{Performance of unseen test sets on Books reviews}
        \label{tab:books}
    \end{minipage}

}}
\end{table*}

\begin{table*}\footnotesize
\centering
\makebox[0pt][c]{\parbox{1\textwidth}{%
    \begin{minipage}[t]{0.45\hsize}\centering
        \begin{tabular}{ llll }
           \hline
          \textbf{Model}  & \textbf{Pr.} & \textbf{Rc.} &\textbf{F-1} \\ \hline
SVM$-$Bigram$-$cor & 	0.83 &	0.82 &	0.82\\
SVM$-$BOWS$-$cor	& 	\textbf{0.84} &	\textbf{0.84} &	\textbf{0.84} \\
SVM$-$Unigram$-$cor	&  0.82 &	0.82 &	0.82 \\
SVM$-$Bigram &	 0.84 &	0.83 &	0.83 \\
SVM$-$BOWS &  0.82 & 0.82 &	0.82 \\
SVM$-$Unigram & 	0.85 &	0.85 &	0.85$\dagger$	\\
SVM$-$Baseline & 	0.70 &	0.70 &	0.70\\ \hline
NB$-$Bigram$-$cor & 	0.84 &	0.84 &	0.84\\
NB$-$BOWS$-$cor	& 0.80  &	0.79 &	0.79 \\
NB$-$Unigram$-$cor	& \textbf{0.82} &	\textbf{0.82} &	\textbf{0.82*} \\
NB$-$Bigram &	0.83 &	0.79 &	0.79 \\
NB$-$BOWS & 0.77 & 0.77 &	0.76 \\
NB$-$Unigram & 0.79 &	0.77 &	0.77\\
NB$-$Baseline & 0.72 &	0.72 &	0.72\\ \hline
        \end{tabular}
        \caption{Performance of unseen test sets on Software reviews}
        \label{tab:software}
    \end{minipage}
    \hspace{\fill}
    \begin{minipage}[t]{0.45\hsize}\centering
        \begin{tabular}{ llll }
           \hline
          \textbf{Model}  & \textbf{Pr.} & \textbf{Rc.} &\textbf{F-1} \\ \hline
SVM$-$Bigram$-$cor & 	0.79 &	0.79 &	0.79\\
SVM$-$BOWS$-$cor	& 	\textbf{0.85} &	\textbf{0.85} &	\textbf{0.85*} \\
SVM$-$Unigram$-$cor	&  0.79 &	0.79 &	0.79 \\
SVM$-$Bigram &	 0.81 &	 0.79&	0.79 \\
SVM$-$BOWS &  0.81 & 0.8 &	0.80 \\
SVM$-$Unigram & 	0.79 &	0.79 &	0.79\\
SVM$-$Baseline & 	0.74 &	0.74 &	0.74\\ \hline
NB$-$Bigram$-$cor & 	\textbf{0.85} &	\textbf{0.82}	 &	\textbf{0.82}\\
NB$-$BOWS$-$cor	& 0.82 &	0.82 &	0.82 \\
NB$-$Unigram$-$cor	& 0.82 &	0.82 &	0.82 \\
NB$-$Bigram &	0.84 &	0.84 &	0.84$\dagger$ \\
NB$-$BOWS & 0.77 & 0.76 &	0.76\\
NB$-$Unigram & 0.82 &	0.82 &	0.82\\
NB$-$Baseline & 0.72 &	0.72 &	0.72\\ \hline

        \end{tabular}
        \caption{Performance of unseen test sets on Kitchen reviews}
        \label{tab:kitchen}
    \end{minipage}

}}
\end{table*}

We see that the model with our correction techniques outperformed the baseline model without the correction techniques on all domains. Other than the baseline model, our technique show comparable performance with the standard bigram, bag-of-words, and unigram models. Not surprisingly, SVM performed better than NB in most cases with bag-of-words and unigram features. On the other hand, NB performed better than SVM with bigram features. The improvement on the baseline technique and the comparable performance on the standard models show the importance of our polarity correction techniques as applicable to sentiment classification. It also emphasizes the fact that using the unseen test sets without sentence-level polarity corrections is likely to lead to mis-classification as a result of inconsistent polarities within each review. Perhaps, it could be beneficial to consider the integration of our polarity correction techniques into an independent sentiment classifier for more accurate sentiment classification.

The limitation of our polarity correction techniques, however, could be in the construction and the performance of the initial `na\"{\i}ve' classifier for performing both the training set and the sentence-level polarity corrections. Also, the classifier needed to be trained on each review domain. At the same time, we emphasize that a moderate classifier - taking a NB classifier as an example - trained with the standard bag-of-word features, gives an average of approximately 72\% F-measure across all domains as observed in our results. Therefore, we believe that the process is likely to have a minimal or negligible effect on the resulting sentiment classifier. As such, in favor of a more efficient classification task, especially on very large datasets, we do not recommend sophisticated classifiers for the initial correction processes. We also like to emphasize that any reasonable sentence-level polarity identification technique,\cite{liu:2012} used in place of the `na\"{\i}ve' classifier in the correction processes, is likely to work just fine and give improved results for the overall sentiment classification task. 

\section{Conclusions}  
\label{section:conclusion}
In this work, we have proposed a training set and sentence-level polarity correction for the sentiment classification task on review documents. We performed experiments on different Amazon product review domains and show that a sentiment classifier with training set and sentence-level polarity corrections, showed improved performance and outperformed a state-of-the-art sentiment classification baseline on all the review domains. Our correction techniques first remove polarity bias from the training set and then inconsistent sentence-level polarities from both training and testing sets. Given the difficulty of the sentiment classification task \cite{liu:2012}, we believe that the improvement shown by the correction technique is promising and could lead to building a more accurate sentiment classifier.
 
In the future, we will integrate the training and sentence-level polarity correction techniques as part of an independent sentiment detection algorithm and perform larger scale experiment on large datasets such as the SNAP Web Data: Amazon reviews dataset\footnote{http://snap.stanford.edu/data/web-Amazon.html}, which was prepared by McAuley and Leskovec.\cite{mcauley2013}


\bibliographystyle{ieeetr}
\bibliography{list}

\begin{thebibliography}{10}

\bibitem{pang:2002}
B.~Pang, L.~Lee, and S.~Vaithyanathan, ``Thumbs up?: sentiment classification
  using machine learning techniques,'' in {\em Proceedings of the ACL-02
  conference on Empirical methods in natural language processing (EMNLP)},
  pp.~79--86, Association for Computational Linguistics, 2002.

\bibitem{vechtomova:2010}
O.~Vechtomova, ``Facet-based opinion retrieval from blogs,'' {\em Information
  Processing \& Management}, vol.~46, no.~1, pp.~71--88, 2010.

\bibitem{liu:2012}
B.~Liu, ``Sentiment analysis and opinion mining,'' {\em Synthesis Lectures on
  Human Language Technologies}, vol.~5, no.~1, pp.~1--167, 2012.

\bibitem{cambria:2013}
E.~Cambria, B.~Schuller, Y.~Xia, and C.~Havasi, ``New avenues in opinion mining
  and sentiment analysis,'' {\em IEEE Intelligent Systems}, vol.~28, no.~2,
  pp.~15--21, 2013.

\bibitem{john_blitzer:2007}
J.~Blitzer, M.~Dredze, and F.~Pereira, ``Biographies, bollywood, boom-boxes and
  blenders: Domain adaptation for sentiment classification,'' (Association of
  Computational Linguistics (ACL)), 2007.

\bibitem{martinez:2014}
E.~Mart{\'\i}nez-C{\'a}mara, M.~T. Mart{\'i}n-Valdivia, L.~A.
  Ure{\~n}a-L{\'o}pez, and A.~R. Montejo-R{\'a}ez, ``Sentiment analysis in
  twitter,'' {\em Natural Language Engineering}, vol.~20, no.~01, pp.~1--28,
  2014.

\bibitem{vilares:2015}
D.~Vilares, M.~A. Alonso, and C.~G{\'o}mez-Rodr{\'i}guez, ``A syntactic
  approach for opinion mining on spanish reviews,'' {\em Natural Language
  Engineering}, pp.~139--163, 2015.

\bibitem{mendel:2010}
J.~Mendel, L.~Zadeh, E.~Trillas, R.~Yager, J.~Lawry, H.~Hagras, and
  S.~Guadarrama, ``What computing with words means to me [discussion forum],''
  {\em Computational Intelligence Magazine, IEEE}, vol.~5, no.~1, pp.~20--26,
  2010.

\bibitem{keshtkar:2012}
F.~Keshtkar and D.~Inkpen, ``A hierarchical approach to mood classification in
  blogs,'' {\em Natural Language Engineering}, vol.~18, no.~01, pp.~61--81,
  2012.

\bibitem{reyes:2014}
A.~Reyes and P.~Rosso, ``On the difficulty of automatically detecting irony:
  beyond a simple case of negation,'' {\em Knowledge and Information Systems},
  vol.~40, no.~3, pp.~595--614, 2014.

\bibitem{melero:2014}
M.~Melero, M.~Costa-Juss{\`a}, P.~Lambert, and M.~Quixal, ``Selection of
  correction candidates for the normalization of spanish user-generated
  content,'' {\em Natural Language Engineering}, pp.~1--27, 2014.

\bibitem{jo:2011}
Y.~Jo and A.~H. Oh, ``Aspect and sentiment unification model for online review
  analysis,'' in {\em Proceedings of the fourth ACM international conference on
  Web search and data mining}, pp.~815--824, ACM, 2011.

\bibitem{tsai:2013}
A.~Tsai, R.~Tsai, and J.~Hsu, ``Building a concept-level sentiment dictionary
  based on commonsense knowledge,'' {\em IEEE Intelligent Systems}, vol.~28,
  no.~2, pp.~22--30, 2013.

\bibitem{poria:2013}
S.~Poria, A.~Gelbukh, A.~Hussain, D.~Das, and S.~Bandyopadhyay, ``Enhanced
  senticnet with affective labels for concept-based opinion mining,'' {\em IEEE
  Intelligent Systems}, vol.~28, no.~2, pp.~31--38, 2013.

\bibitem{roman:2015}
N.~T. Roman, P.~Piwek, A.~M. B.~R. Carvalho, and A.~R. Alvares, ``Sentiment and
  behaviour annotation in a corpus of dialogue summaries,'' {\em Journal of
  Universal Computer Science}, vol.~21, no.~4, pp.~561--586, 2015.

\bibitem{pang:2008}
B.~Pang and L.~Lee, ``Opinion mining and sentiment analysis,'' {\em Foundations
  and Trends in Information Retrieval}, vol.~2, no.~1-2, pp.~1--135, 2008.

\bibitem{devitt:2007}
A.~Devitt and K.~Ahmad, ``Sentiment polarity identification in financial news:
  A cohesion-based approach,'' in {\em Proceedings of the 45th Annual Meeting
  of the Association of Computational Linguistics}, pp.~984--991, ACL, 2007.

\bibitem{brabazon:2008}
A.~Brabazon, M.~O'Neill, and I.~Dempsey, ``An introduction to evolutionary
  computation in finance,'' {\em Computational Intelligence Magazine, IEEE},
  vol.~3, no.~4, pp.~42--55, 2008.

\bibitem{fortuny:2014}
E.~J. Fortuny, T.~D. Smedt, D.~Martens, and W.~Daelemans, ``Evaluating and
  understanding text-based stock price prediction models,'' {\em Information
  Processing \& Management}, vol.~50, no.~2, pp.~426 -- 441, 2014.

\bibitem{li:2010extraction}
D.~Li, A.~Laurent, P.~Poncelet, and M.~Roche, ``Extraction of unexpected
  sentences: A sentiment classification assessed approach,'' {\em Intelligent
  Data Analysis}, vol.~14, no.~1, p.~31, 2010.

\bibitem{jia:2009}
L.~Jia, C.~Yu, and W.~Meng, ``The effect of negation on sentiment analysis and
  retrieval effectiveness,'' in {\em Proceeding of the 18th ACM conference on
  Information and knowledge management}, (Hong Kong, China), pp.~1827--1830,
  ACM, 2009.

\bibitem{pang:2004}
B.~Pang and L.~Lee, ``A sentimental education: sentiment analysis using
  subjectivity summarization based on minimum cuts,'' in {\em Proceedings of
  the 42nd Annual Meeting on Association for Computational Linguistics},
  (Barcelona, Spain), p.~271, Association for Computational Linguistics, 2004.

\bibitem{wilson:2005a}
T.~Wilson, J.~Wiebe, and P.~Hoffmann, ``Recognizing contextual polarity in
  phrase-level sentiment analysis,'' in {\em Proceedings of the conference on
  Human Language Technology and Empirical Methods in Natural Language
  Processing}, (Vancouver, British Columbia, Canada), pp.~347--354, Association
  for Computational Linguistics, 2005.

\bibitem{choi:2008}
Y.~Choi and C.~Cardie, ``Learning with compositional semantics as structural
  inference for subsentential sentiment analysis,'' in {\em Proceedings of the
  Conference on Empirical Methods in Natural Language Processing}, (Honolulu,
  Hawaii), pp.~793--801, Association for Computational Linguistics, 2008.

\bibitem{wiebe:2005}
J.~Wiebe, T.~Wilson, and C.~Cardie, ``Annotating expressions of opinions and
  emotions in language,'' {\em Language Resources and Evaluation}, vol.~39,
  no.~2/3, pp.~165--210, 2005.

\bibitem{bautin:2010}
M.~Bautin, C.~B. Ward, A.~Patil, and S.~S. Skiena, ``Access: news and blog
  analysis for the social sciences,'' in {\em Proceedings of the 19th
  international conference on World wide web}, (Raleigh, North Carolina, USA),
  pp.~1229--1232, ACM, 2010.

\bibitem{lee:2010a}
Y.~Lee, H.-y. Jung, W.~Song, and J.-H. Lee, ``Mining the blogosphere for top
  news stories identification,'' in {\em Proceeding of the 33rd international
  ACM SIGIR conference on Research and development in information retrieval},
  (Geneva, Switzerland), pp.~395--402, ACM, 2010.

\bibitem{tang:2009}
H.~Tang, S.~Tan, and X.~Cheng, ``A survey on sentiment detection of reviews,''
  {\em Expert Systems with Applications}, vol.~36, no.~7, pp.~10760--10773,
  2009.

\bibitem{arel:2010}
I.~Arel, D.~C. Rose, and T.~P. Karnowski, ``Deep machine learning-a new
  frontier in artificial intelligence research [research frontier],'' {\em
  Computational Intelligence Magazine, IEEE}, vol.~5, no.~4, pp.~13--18, 2010.

\bibitem{he:2011}
Y.~He and D.~Zhou, ``Self-training from labeled features for sentiment
  analysis,'' {\em Information Processing \& Management}, vol.~47, no.~4,
  pp.~606--616, 2011.

\bibitem{calais:2011}
P.~H. Calais~Guerra, A.~Veloso, W.~Meira~Jr, and V.~Almeida, ``From bias to
  opinion: a transfer-learning approach to real-time sentiment analysis,'' in
  {\em Proceedings of the 17th ACM SIGKDD international conference on Knowledge
  discovery and data mining}, pp.~150--158, ACM, 2011.

\bibitem{hang_cui:2006}
H.~Cui, V.~Mittal, and M.~Datar, ``Comparative experiments on sentiment
  classification for online product reviews,'' (American Association for
  Artificial Intelligence (AAAI)), 2006.

\bibitem{maldonado:2014}
S.~Maldonado and C.~Montecinos, ``Robust classification of imbalanced data
  using one-class and two-class svm-based multiclassifiers,'' {\em Intelligent
  Data Analysis}, vol.~18, no.~1, pp.~95--112, 2014.

\bibitem{tan:2009}
S.~Tan, X.~Cheng, Y.~Wang, and H.~Xu, ``Adapting naive bayes to domain
  adaptation for sentiment analysis,'' in {\em Advances in Information
  Retrieval}, pp.~337--349, Springer, 2009.

\bibitem{ye:2009}
Q.~Ye, Z.~Zhang, and R.~Law, ``Sentiment classification of online reviews to
  travel destinations by supervised machine learning approaches,'' {\em Expert
  Systems with Applications}, vol.~36, no.~3, pp.~6527--6535, 2009.

\bibitem{xia:2013}
R.~Xia, C.~Zong, X.~Hu, and E.~Cambria, ``Feature ensemble plus sample
  selection: A comprehensive approach to domain adaptation for sentiment
  classification,'' {\em IEEE Intelligent Systems}, vol.~28, no.~3, pp.~10--18,
  2013.

\bibitem{lowd:2005}
D.~Lowd and P.~Domingos, ``Naive bayes models for probability estimation,'' in
  {\em Proceedings of the 22nd international conference on Machine learning},
  ICML '05, (New York, NY, USA), pp.~529--536, ACM, 2005.

\bibitem{tan:2011}
L.~Tan, J.~Na, Y.~Theng, and K.~Chang, ``Sentence-level sentiment polarity
  classification using a linguistic approach,'' {\em Digital Libraries: For
  Cultural Heritage, Knowledge Dissemination, and Future Creation}, pp.~77--87,
  2011.

\bibitem{wilson:2009}
T.~Wilson, J.~Wiebe, and P.~Hoffmann, ``Recognizing contextual polarity: An
  exploration of features for phrase-level sentiment analysis,'' {\em
  Computational Linguistics}, vol.~35, no.~3, pp.~399--433, 2009.

\bibitem{li:2010}
S.~Li, S.~Y.~M. Lee, Y.~Chen, C.-R. Huang, and G.~Zhou, ``Sentiment
  classification and polarity shifting,'' in {\em Proceedings of the 23rd
  International Conference on Computational Linguistics}, (Beijing, China),
  pp.~635--643, Association for Computational Linguistics, 2010.

\bibitem{platt:1998}
J.~Platt, ``Sequential minimal optimization: A fast algorithm for training
  support vector machines,'' Tech. Rep. MSR-TR-98-14, Microsoft Research, 1998.

\bibitem{rish:2001}
I.~Rish, ``An empirical study of the naive bayes classifier,'' in {\em IJCAI
  2001 workshop on empirical methods in artificial intelligence}, vol.~3,
  pp.~41--46, 2001.

\bibitem{hall:2009}
M.~Hall, E.~Frank, G.~Holmes, B.~Pfahringer, P.~Reutemann, and I.~H. Witten,
  ``The weka data mining software: an update,'' {\em ACM SIGKDD explorations
  newsletter}, vol.~11, no.~1, pp.~10--18, 2009.

\bibitem{thornton:2013}
C.~Thornton, F.~Hutter, H.~H. Hoos, and K.~Leyton-Brown, ``Auto-weka: Combined
  selection and hyperparameter optimization of classification algorithms,'' in
  {\em Proceedings of the 19th ACM SIGKDD international conference on Knowledge
  discovery and data mining}, pp.~847--855, ACM, 2013.

\bibitem{mcauley2013}
J.~McAuley and J.~Leskovec, ``Hidden factors and hidden topics: understanding
  rating dimensions with review text,'' in {\em Proceedings of the 7th ACM
  conference on Recommender systems}, pp.~165--172, ACM, 2013.

\end{thebibliography}

\end{document}